\documentclass{article}

\usepackage[T1]{fontenc}
\usepackage{lmodern}% scalable Latin Modern (text)
\usepackage{type1cm}% scalable Computer Modern (incl. math fonts at arbitrary sizes)
\usepackage{microtype}
\usepackage{graphicx}
\usepackage{booktabs}
\PassOptionsToPackage{table}{xcolor}% icml2027 loads xcolor; enable table options
\usepackage[hyperfootnotes=false]{hyperref}
\usepackage[accepted]{icml2027}
\usepackage{amsmath,amssymb,amsthm}
\usepackage{placeins}
\usepackage{setspace}
\definecolor{hlrowcolor}{RGB}{224,236,246}
\newtheorem{proposition}{Proposition}

\icmltitlerunning{FN-Prune: Finite RMSNorm Perturbations}
\begin{document}
\twocolumn[
\icmltitle{FN-Prune: Training-Free Expert Pruning with Finite RMSNorm Perturbations}
\begin{icmlauthorlist}
\icmlauthor{Zeyan Li}{sjtu}
\icmlauthor{Jianfeng Xu}{sjtu}
\end{icmlauthorlist}
\icmlaffiliation{sjtu}{Shanghai Jiao Tong University, Shanghai, China}
\icmlcorrespondingauthor{Zeyan Li}{zeyan0823@sjtu.edu.cn}
\icmlkeywords{Mixture-of-experts, model compression, expert pruning, RMSNorm, training-free pruning}
\vskip 0.3in
]
\printAffiliationsAndNotice{}
\begin{abstract}
Retaining model quality when pruning mixture-of-experts language models depends on which experts are removed. Frequency- and magnitude-based scores measure the expert in isolation and overlook how normalization consumes its contribution, since subtracting one expert changes both the direction of the residual stream and the scale applied to what remains. We propose FN-Prune, a training-free criterion that measures this cost directly. Under the checkpoint's original routing, it subtracts each router-weighted expert contribution and scores the induced squared change in the next RMSNorm output, averaged over all calibration tokens in a single forward pass. Low-scoring experts are then physically deleted at uniform per-layer ratios, roughly halving the materialized footprint at $50\%$ removal while raising generation throughput. At that budget, FN-Prune achieves lower WikiText-2 perplexity than the best published methods on Qwen3, ERNIE, and OLMoE, with a $32\%$ reduction on OLMoE, and improves GSM8K accuracy over the best published methods by $7.5$ and $7.0$ percentage points on Qwen3 and ERNIE. Across three model families and two removal budgets, these results establish the finite normalized perturbation as an effective training-free basis for expert pruning.
\end{abstract}
\section{Introduction}
\label{sec:intro}

Sparse mixture-of-experts (MoE) language models activate only a few experts per token, decoupling parameter count from per-token compute~\citep{lepikhin2021gshard,fedus2022switch,du2022glam,dai2024deepseekmoe,jiang2024mixtral,deepseekai2024v3,muennighoff2024olmoe,qwen2024technical}. The sparsity does not remove the weights of inactive experts from the model, and physical expert pruning addresses that memory cost. Removing half of Qwen3's routed experts reduces its materialized parameter-plus-buffer footprint from $56.87$ to $29.86$ GiB. The difficult decision is which experts to remove while retaining useful behavior without recovery training.

\begin{figure}[!t]
\centering
\includegraphics[width=\columnwidth]{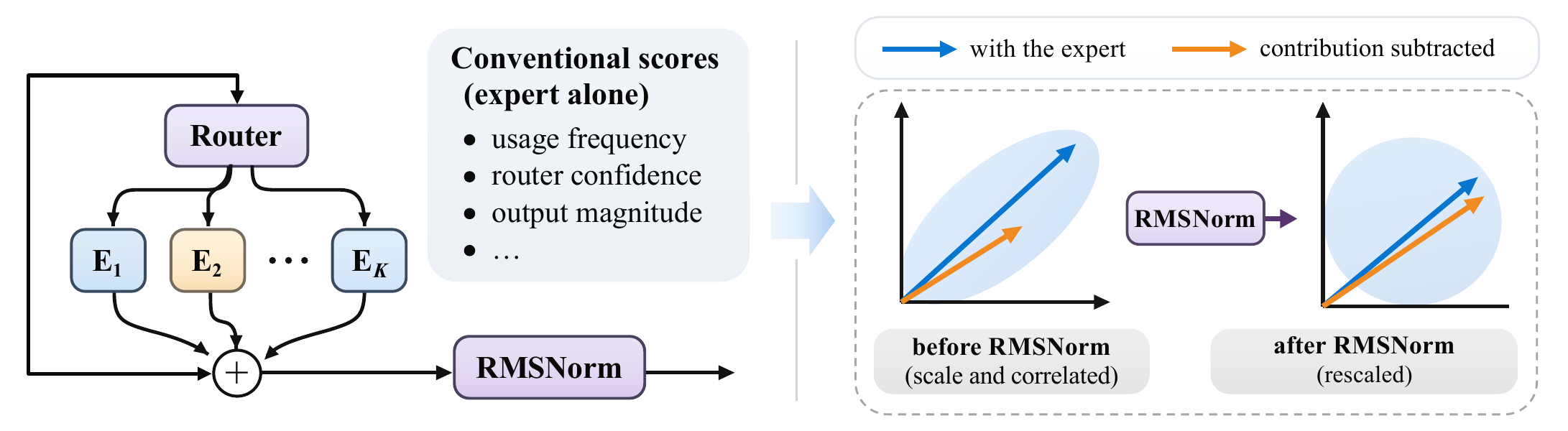}
\caption{\textbf{Scoring an expert by what the network reads.} FN-Prune occludes one weighted contribution under fixed routing and compares the next RMSNorm output before and after; aligned contributions are largely absorbed by the renormalized scale.}
\label{fig:intro_teaser}
\end{figure}

Any such decision implicitly ranks experts by how much their removal would cost the model. Existing rankings estimate that cost from the expert alone. Usage frequency counts selections~\citep{lu2024notallequal,zhou2022mixture,huang2025remoe}, gate mass weights them, and output norms describe what the expert computes~\citep{li2023mergemsmoe,liu2025simbal}. A rarely selected expert can carry substantial gate mass, and a frequently selected one can produce small outputs, yet all of these signals are measured upstream of the operation that actually consumes the contribution. In the blocks studied here, that consumer is normalization. Each expert output is added to a residual stream that the next RMSNorm rescales. Removing one contribution therefore changes both the direction of that stream and the scale applied to everything left in it, even when the other contributions are unchanged. A contribution aligned with the residual mostly changes a magnitude that normalization absorbs; an equally large contribution in another direction changes where the model looks (Fig.~\ref{fig:intro_teaser}). The cost of removing an expert is thus contextual, jointly determined by its contribution and the residual state that contribution enters, and is not readable from the contribution's own frequency or magnitude.

Motivated by these observations, we propose \textbf{FN-Prune} (Finite RMSNorm-Perturbation Pruning), which measures this cost directly. Holding the checkpoint's original routes and weights fixed, we subtract one expert's native, router-weighted contribution and average the squared change in the next RMSNorm output over the complete calibration corpus:
\begin{equation}
S_\ell(i)\;=\;\mathbb{E}_{t\sim\mathcal D}\Bigl[\bigl\|\Phi(r_t)-\Phi\bigl(r_t-c_i(t)\bigr)\bigr\|^2\Bigr],
\label{eq:intro_score}
\end{equation}
where $\Phi$ is the input RMSNorm of the following block or the final RMSNorm after the last block, $r_t$ is the residual-stream state entering that normalization, and $c_i(t)$ is expert $i$'s weighted contribution under the checkpoint's unmodified routing. Evaluating both normalized states retains the finite change in scale after subtraction. Averaging over the whole population makes the score the product of how often an expert is used and how much its removal disturbs the normalized state when it is used. Both states come from the same unmodified forward pass, so one calibration pass produces every score; low-scoring experts are then physically deleted under exact uniform per-layer quotas, and inference reroutes among the survivors.

We evaluate FN-Prune in two experiments. The first swaps the ranking for frequency deletion under identical quotas and procedure. OLMoE's WikiText-2 perplexity at $50\%$ removal then rises from $32.48$ to $49.14$, with smaller but consistent gaps on Qwen3 and ERNIE, so the gain comes from \emph{which} experts leave. The second compares FN-Prune with ten published compression systems spanning deletion, non-uniform allocation, novice replacement, merging, and recovery training. At $50\%$ removal, FN-Prune attains the lowest WikiText-2 perplexity on every family, including a $32\%$ reduction relative to the best published method on OLMoE, and improves GSM8K over the strongest published method by $7.5$ and $7.0$ percentage points on Qwen3 and ERNIE. In summary, our contributions are as follows.
\begin{enumerate}
\item We propose \textbf{FN-Prune}, a training-free pruning criterion and procedure in which experts are scored by the population expected squared change of the next RMSNorm output under fixed routing, all scores accumulate in one forward pass, and low-scoring experts are deleted under uniform per-layer quotas.
\item We prove a \textbf{decomposition} of the finite perturbation into a removed-content term and an induced scale correction, making explicit why equal-magnitude contributions can have very different removal costs.
\item We demonstrate that the proposed method \textbf{performs well}. At $50\%$ removal, FN-Prune attains the lowest WikiText-2 perplexity among ten published compression systems on every model family.
\end{enumerate}

\section{Related Work}
\label{sec:related}

Compressing mixture-of-experts models clusters along three axes, namely the signal for ranking experts, the reuse of retained capacity, and the allocation of removals across layers. The subsections follow this order.

\subsection{Ranking Experts for Removal}
\label{sec:related_within_layer}

How often an expert is selected predicts removal damage well enough that frequency pruning yields smooth quality curves at moderate keep ratios~\citep{lu2024notallequal}, and router probabilities and gate activations refine the same participation signal~\citep{zhou2022mixture,huang2025remoe}. A second family looks at what the expert computes. Weighting output norms by router gates targets MoE-layer reconstruction error~\citep{lasby2025reap}, and similarity between experts indicates which ones can be consolidated~\citep{li2023mergemsmoe,liu2025simbal}. Both families, however, measure the expert in isolation, through its participation statistics or the magnitude of its output. The criterion of this paper scores the change that a removal induces in the normalized state the next block actually reads, which brings the residual context and the recomputed scale into the ranking.

A more recent line is explicitly residual-aware. Each calibration token is weighted by how much the routed block changes the residual direction, and experts are scored under that weighting in one pass from a small in-domain calibration set~\citep{dong2025easyep}. That token-level weight is shared by every expert the token selects, whereas the finite comparison used here is specific to the expert whose contribution is removed and is averaged over the complete calibration corpus. Removal decisions have also borrowed from reconstruction-based dense pruning~\citep{frantar2022obc} and from subset search, whether by evolutionary optimization over deletion plans~\citep{liu2024eep} or by marginal-contribution estimation through Monte Carlo coalition evaluation with early truncation and router-guided sampling~\citep{huang2025shapley}; the latter's preservation guarantee assumes additive utility and bounded estimation error, and all of these routes require repeated evaluations or proxy objectives, while both states scored here come from a single fixed-routing forward pass. Selection can also protect specific capabilities, safety in particular~\citep{lai2025safex,kim2025safemoe}; parallel work characterizes expert roles through task, semantic, and geometric lenses~\citep{olson2025semanticrouting,herbst2026expertstrikesback,ternovtsii2026geometricrouting,ye2026polysemanticpaths}.

\subsection{Reusing and Retraining Capacity}
\label{sec:related_consolidation}

Another family of methods reuses expert capacity. Merging consolidates similar experts~\citep{li2023mergemsmoe}, absorption transfers routing mass between them~\citep{aquino2026absorption}, and related operations preserve pairwise similarity during pruning~\citep{liu2025simbal} or expose redundancy through specialization~\citep{chen2025advancing}. Replacing experts by learned constant modules~\citep{zhang2026mone} and adjusting merged models with recovery training change what the retained parameters compute. We compare these operations at matched routed-expert budgets; the frequency control isolates the ranking's own contribution.

\subsection{Distributing Removals across Layers}
\label{sec:related_allocation}

How many experts each layer can spare is independent of which ones. In dense-model pruning, reconstruction-optimal updates~\citep{frantar2023sparsegpt}, activation-weighted magnitude~\citep{sun2023wanda}, and structured group pruning~\citep{ma2023llmpruner} operate under shared ratios, whereas allocation by outlier structure~\citep{yin2024owl}, by heavy-tailed weight statistics~\citep{lu2024alphapruning}, or learned jointly with width and depth~\citep{xia2023shearedllama} redistributes them. MoE models behave the same way in our preliminary measurements. Reallocating matched removals changes perplexity substantially (Section~\ref{sec:method_motivation}).

\section{FN-Prune}
\label{sec:method}

\subsection{Method Overview}
\label{sec:method_overview}

\begin{figure*}[!tp]
\centering
\includegraphics[width=\textwidth]{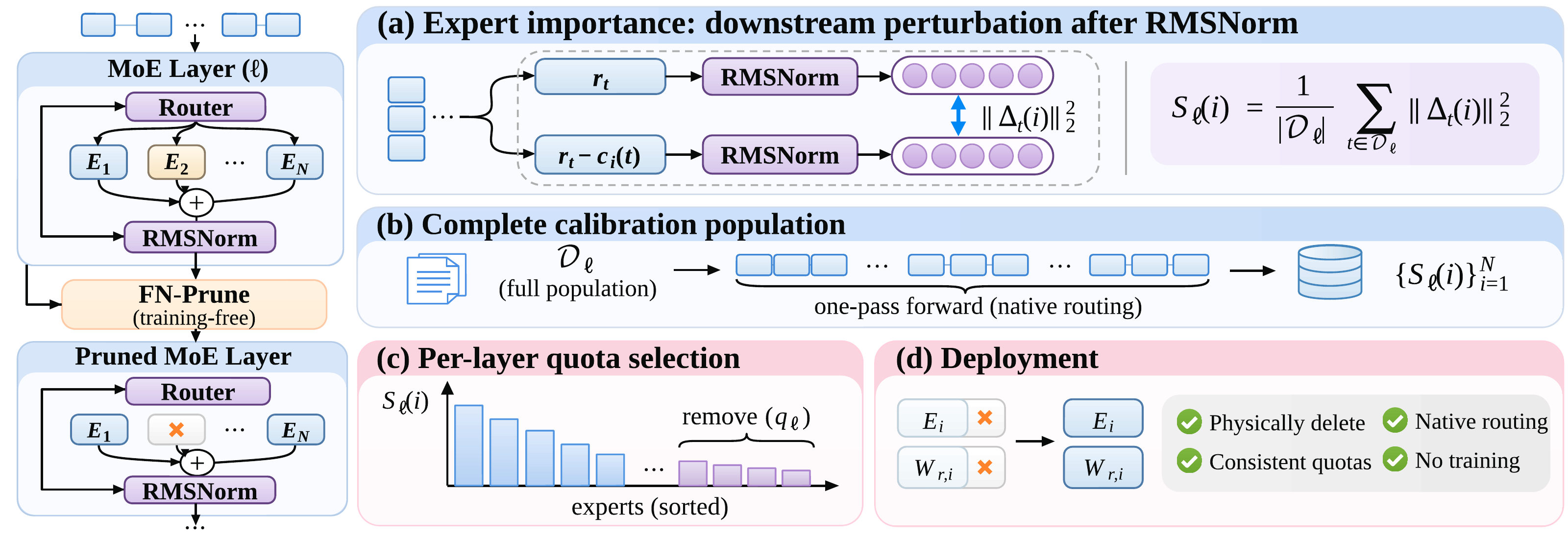}
\caption{\textbf{FN-Prune pipeline.} One native-routing calibration pass scores every expert; each layer deletes its lowest-scoring experts up to its quota, and inference reroutes among the survivors (Algorithm~\ref{alg:fnprune}).}
\label{fig:fnprune_pipeline}
\end{figure*}

FN-Prune ranks experts by the change their removal would induce at the next RMSNorm. For the same token and native routing, we subtract one expert's weighted contribution from the residual and evaluate RMSNorm at both states. Averaging the squared differences over the calibration corpus gives one score per expert; the lowest-scoring experts are then deleted under exact uniform per-layer quotas and inference reroutes among the survivors (Figure~\ref{fig:fnprune_pipeline}).

\subsection{Finite Normalization Score}
\label{sec:method_setup}

Consider a pretrained MoE with layers $\ell\in[L]$, $N_\ell$ routed experts per layer, and top-$k_\ell$ selection. Shared always-on experts, where present, are retained. Vectors lie in $\mathbb{R}^d$; $\odot$ denotes elementwise multiplication, and $\langle\cdot,\cdot\rangle$ and $\|\cdot\|$ are the Euclidean inner product and norm. We write $[n]=\{1,\ldots,n\}$ and omit $\ell$ on quantities defined within one layer.

For token $t$, native routing selects $\mathcal K_\ell(t)$ and assigns weights $w_i(t)$ under the checkpoint's own selected-weight renormalization policy. If $x_t$ is the expert input and $f_i$ the expert function, the routed output is
\begin{equation}
y_\ell(t)=\sum_{i\in\mathcal K_\ell(t)} w_i(t)\, f_i(x_t).
\label{eq:moe_output}
\end{equation}
Let $h_t$ denote the residual before adding that routed output, including any shared-expert output. The next normalization reads
\begin{equation}
r_t \;=\; h_t + y_\ell(t),
\qquad
\Phi(r) \;=\; \gamma \odot \frac{r}{\sqrt{\tfrac1d\|r\|^2+\epsilon}},
\label{eq:residual_rmsnorm}
\end{equation}
Here $\Phi$ is the following block's input RMSNorm, or the final RMSNorm after the last MoE block; $\gamma\in\mathbb{R}^d$ is its gain and $\epsilon>0$ its normalization constant. The contribution to be tested is
\begin{equation}
c_i(t) =
\begin{cases}
w_i(t) f_i(x_t), & i\in\mathcal K_\ell(t),\\
0, & i\notin\mathcal K_\ell(t).
\end{cases}
\label{eq:contribution}
\end{equation}
The average is taken over a fixed calibration population. Calibration uses the complete WikiText-2 training split, tokenized and packed at context length $2{,}048$, retaining the final, possibly partial chunk. During scoring, the unmodified checkpoint runs in evaluation mode on the fixed, fully packed calibration stream; we preserve its native selection rule, routing weights, and any correction bias (a learned per-expert routing offset in ERNIE's gate). The population $\mathcal D_\ell$ contains every non-padding token processed at layer $\ell$.

\paragraph{Population score}
\label{sec:method_score}

With routes and weights fixed, removing $c_i(t)$ changes the normalized residual from $\Phi(r_t)$ to $\Phi(r_t-c_i(t))$. FN-Prune averages the squared norm of this finite change:
\begin{equation}
S_\ell(i)
\;=\;
\frac{1}{|\mathcal D_\ell|}
\sum_{t\in\mathcal D_\ell}
\bigl\|
\Phi(r_t)-\Phi\bigl(r_t-c_i(t)\bigr)
\bigr\|^2,
\label{eq:fn_score}
\end{equation}
The denominator is the layer's entire calibration population, $T_\ell=|\mathcal D_\ell|$. Because unselected tokens contribute zero, the score equals the expert's selection frequency times its conditional mean squared perturbation when selected.

\paragraph{Finite normalization effect}
\label{sec:method_properties}

The finite comparison involves both the removed vector and the normalization scale of the remaining residual.

\begin{proposition}[Consumer-side decomposition]
\label{prop:decomposition}
For any $r,c\in\mathbb{R}^d$, writing $v=\tfrac1d\|r\|^2+\epsilon$ and $v'=\tfrac1d\|r-c\|^2+\epsilon$,
\begin{equation}
\Phi(r)-\Phi(r-c)
\;=\;
\gamma\odot\Biggl[
\frac{c}{\sqrt{v'}}
\;-\;
r\,\frac{2\langle r,c\rangle-\|c\|^2}{d\bigl(\sqrt{v}+\sqrt{v'}\bigr)\sqrt{v}\sqrt{v'}}
\Biggr].
\label{eq:fn_decomposition}
\end{equation}
\end{proposition}
The proof is given in Appendix~\ref{app:proofs}. Before the channelwise gain $\gamma$ is applied, the first term is a scaled copy of the removed contribution and the second is a radial correction parallel to the residual, generated by the change in normalization scale. The two terms can reinforce or cancel; the radial coefficient vanishes when $2\langle r,c\rangle=\|c\|^2$, and the gain $\gamma$ reweights both terms channelwise. Two contributions of equal norm can therefore induce very different normalized perturbations, depending on their relation to the residual.

A natural alternative is the linearization of the same comparison. The exact RMSNorm derivative gives its infinitesimal version,
\begin{equation}
J_\Phi(r)[c]
=
\gamma\odot
\frac{c - r\,\dfrac{\langle r,c\rangle/d}{\tfrac1d\|r\|^2+\epsilon}}
{\sqrt{\tfrac1d\|r\|^2+\epsilon}},
\label{eq:rmsnorm_jvp}
\end{equation}
and Section~\ref{sec:method_controls} compares this local control against the finite difference, routed energy, and a scale-only alternative.

\subsection{Expert Selection and Physical Pruning}
\label{sec:method_selection}

For budget $b\in\{0.25,0.5\}$, each layer removes the same fraction of its routed experts. The quota and selected set are
\begin{equation}
q_\ell=\lfloor N_\ell\, b\rfloor,
\qquad
\mathcal R_\ell
=
\operatorname*{arg\,min}_{\substack{\mathcal R\subseteq[N_\ell]\\|\mathcal R|=q_\ell}}
\sum_{i\in\mathcal R} S_\ell(i),
\label{eq:quota_selection}
\end{equation}
Ties are broken by expert index. For the evaluated architectures and budgets, $N_\ell b$ is an integer and $q_\ell\le N_\ell-k_\ell$, so the realized global count is exactly $\sum_\ell N_\ell b$ and every layer retains enough experts for native top-$k_\ell$ selection.

Deletion acts on the checkpoint itself. We remove the selected expert modules, slice the corresponding router-projection rows, update routed-expert counts, and slice any matching correction-bias entries in ERNIE. Shared experts are unchanged. Releasing removed weights lowers the parameter-plus-buffer footprint. Algorithm~\ref{alg:fnprune} gives the full procedure, including the coverage checks and the tie rule.

\begin{algorithm}[!t]
\caption{One-pass scoring and physical deletion}
\label{alg:fnprune}
\begin{algorithmic}[1]
\REQUIRE Pretrained MoE; calibration stream $\mathcal D$; budget $b$.
\ENSURE Physically pruned model; deletion record $\{\mathcal R_\ell\}$.
\STATE Instrument routers, experts, and next-layer RMSNorms.
\FOR{each batch of $\mathcal D$ with padding excluded}
\STATE Record native $\mathcal K_\ell(t)$, routing weights $w_i(t)$, expert outputs, and $r_t$.
\STATE For each selected $(\ell,i,t)$, evaluate and accumulate the squared map difference (the summand of Eq.~\ref{eq:fn_score}).
\ENDFOR
\STATE Verify population, top-$k$ coverage, and finiteness.
\FOR{$\ell=1,\ldots,L$}
\STATE Divide each expert's sum by the token count $T_\ell$.
\STATE Set $q_\ell=\lfloor N_\ell b\rfloor$. Select the $q_\ell$ lowest means for $\mathcal R_\ell$, breaking ties by expert index.
\ENDFOR
\STATE Remove the selected experts, their router rows and any correction-bias entries. Update the expert counts.
\STATE \textbf{return} pruned model, $\{\mathcal R_\ell\}$.
\end{algorithmic}
\end{algorithm}

\subsection{Native Rerouting and Computational Cost}
\label{sec:method_routing_boundary}

The scoring intervention removes one contribution while the routes and weights stay fixed. Deployment changes both, since a physically pruned model re-selects top-$k_\ell$ experts from the survivors and applies its native weight policy. We isolate the weight-renormalization component first. Consider a selected expert $i$ for a token whose selected weights sum to one, with $w_i(t)<1$; remove $i$ while holding the selected set at $\mathcal K_\ell(t)\setminus\{i\}$ and rescale the surviving weights to sum to one.

\begin{proposition}[Survivor renormalization]
\label{prop:refill}
Under this construction, the change in the residual stream is
\begin{equation}
\begin{aligned}
\Delta r_t
&=\;
\frac{w_i(t)\, y_\ell(t)-c_i(t)}{1-w_i(t)}
\\
&=\;
\frac{w_i(t)}{1-w_i(t)}\,\bigl(y_\ell(t)-f_i(x_t)\bigr),
\end{aligned}
\label{eq:refill}
\end{equation}
whereas the fixed-routing subtraction in Eq.~\ref{eq:fn_score} changes the residual by $-c_i(t)$.
\end{proposition}
The proof is given in Appendix~\ref{app:proofs}. Equation~\ref{eq:refill} isolates survivor rescaling; deployment also refills the top-$k$ set.

\paragraph{One-pass accumulation}
\label{sec:method_complexity}

Both states required by the score are available from the unmodified calibration forward, namely the residual entering the next RMSNorm and the selected expert's weighted output, so the entire score accumulates in one pass.

\begin{proposition}[One-pass accumulation]
\label{prop:onepass}
With native routing fixed over the complete processed calibration population, $S_\ell(i)$ is the mean squared norm of Eq.~\ref{eq:fn_decomposition} and can be accumulated in one forward pass over $\mathcal D$. Evaluating the difference from the cached pair $(r_t,c_i(t))$ takes $O(d)$ arithmetic per selected token--expert pair. The accumulators use $O(\sum_\ell N_\ell)$ scalars.
\end{proposition}
The proof is given in Appendix~\ref{app:proofs}. An OLMoE-1B-7B diagnostic compares each calibration prefix with the complete-pass ranking. The median layer reaches a Spearman rank correlation of $0.99$ after about $0.23$M input tokens, and all sixteen layers sustain at least $0.99$ from about $1.49$M tokens through the end of the $2.44$M-token pass. The deployed score is computed over the complete calibration population.

\section{Experiments}
\label{sec:campaign}
\label{sec:results}

Our evaluation is organized around four questions:
\begin{itemize}
\setlength{\itemsep}{1pt}
\setlength{\parskip}{0pt}
\setlength{\topsep}{3pt}
    \item \textbf{RQ1 (Published baselines).} How does FN-Prune compare with published methods at matched budgets?
    \item \textbf{RQ2 (Design choices).} Which scoring, aggregation and allocation choices matter, and does the selected score beat routing frequency under matched deletion?
    \item \textbf{RQ3 (Capability retention).} Which downstream capabilities are retained after pruning?
    \item \textbf{RQ4 (Physical efficiency).} Do physical deletions reduce model size and improve throughput?
\end{itemize}

% Queue the full-width RQ1 table at the section head so it can occupy the
% next page top (the intended reading position, next to the RQ1 narrative).
\begingroup\singlespacing
% Generated by tools/render_baseline_paper_tables_v2.py; do not edit numeric rows.
% Source: evidence/baseline_results_final_96_20261005/baseline_campaign_audit.json
\begin{table*}[!tb]
\centering
\caption{Per-task results for FN-Prune (shaded) and the ten published methods; each metric pairs the 25\% and 50\% removal budgets, with the full model as unpruned reference.}
\label{tab:baseline_results}
\fontsize{7.2}{8.5}\selectfont
\setlength{\tabcolsep}{0.8pt}
\renewcommand{\arraystretch}{1.06}
\begin{tabular*}{\textwidth}{@{\extracolsep{\fill}}lrrrrrrrrrr@{}}
\toprule
& \multicolumn{2}{c}{MMLU} & \multicolumn{2}{c}{OBQA} & \multicolumn{2}{c}{WinoG.} & \multicolumn{2}{c}{WT2 PPL} & \multicolumn{2}{c}{GSM8K} \\
\cmidrule(lr){2-3}\cmidrule(lr){4-5}\cmidrule(lr){6-7}\cmidrule(lr){8-9}\cmidrule(lr){10-11}
Method & 25\% & 50\% & 25\% & 50\% & 25\% & 50\% & 25\% & 50\% & 25\% & 50\% \\
\midrule
\multicolumn{11}{@{}l}{\textbf{\textit{Qwen3-30B-A3B}}} \\
Full model & \multicolumn{2}{c}{77.85} & \multicolumn{2}{c}{45.20} & \multicolumn{2}{c}{70.48} & \multicolumn{2}{c}{9.14} & \multicolumn{2}{c}{88.40} \\
\noalign{\vskip 1pt\hbox to \textwidth{\xleaders\hbox{\rule{2.2pt}{0.25pt}\hskip 2pt}\hfill}\vskip 1pt}
M-SMoE~{\fontsize{6.2}{6.8}\selectfont\citeyearpar{li2023mergemsmoe}} & 72.87 & 46.44 & 44.80 & 40.00 & 70.64 & 68.03 & 9.25 & 11.49 & 82.64 & 51.18 \\
MC-SMoE~{\fontsize{6.2}{6.8}\selectfont\citeyearpar{li2023mergemsmoe}} & 72.88 & 46.22 & \underline{45.20} & 39.80 & \underline{71.19} & \textbf{69.30} & \textbf{9.21} & \underline{10.35} & 82.64 & 55.57 \\
OWL-to-MoE~{\fontsize{6.2}{6.8}\selectfont\citeyearpar{yin2024owl}} & 74.19 & 55.95 & 43.20 & 40.20 & 69.93 & \underline{68.90} & 9.31 & 10.60 & 81.80 & 26.23 \\
NAEE~{\fontsize{6.2}{6.8}\selectfont\citeyearpar{lu2024notallequal}} & \underline{74.65} & 47.12 & 44.00 & 36.60 & 70.96 & 67.40 & 9.35 & 12.81 & 82.94 & \underline{62.02} \\
EEP-Prune~{\fontsize{6.2}{6.8}\selectfont\citeyearpar{liu2024eep}} & \textbf{75.03} & 60.32 & 43.80 & \textbf{42.20} & 70.56 & 67.56 & 9.32 & 10.64 & 82.71 & 56.10 \\
AlphaPruning-to-MoE~{\fontsize{6.2}{6.8}\selectfont\citeyearpar{lu2024alphapruning}} & 65.05 & 23.48 & 37.40 & 25.20 & 64.25 & 51.07 & 18.17 & 149.52 & 6.52 & 0.00 \\
HC-SMoE~{\fontsize{6.2}{6.8}\selectfont\citeyearpar{chen2025hcsmoe}} & 63.73 & 41.16 & 38.00 & 27.00 & 64.40 & 56.43 & 13.92 & 32.78 & \underline{86.81} & 59.59 \\
MoE Pathfinder~{\fontsize{6.2}{6.8}\selectfont\citeyearpar{yang2025moepathfinder}} & 73.62 & \underline{61.05} & 44.20 & 40.80 & 69.53 & 67.96 & \underline{9.23} & 10.50 & \textbf{87.41} & 44.12 \\
REAP-Prune~{\fontsize{6.2}{6.8}\selectfont\citeyearpar{lasby2025reap}} & 40.33 & 23.14 & 32.20 & 26.80 & 57.22 & 51.46 & 20.58 & 60.99 & 79.23 & 21.91 \\
MoNE~{\fontsize{6.2}{6.8}\selectfont\citeyearpar{zhang2026mone}} & 74.41 & 40.52 & \textbf{45.40} & 31.80 & 70.32 & 62.04 & 9.36 & 17.77 & 79.98 & 49.81 \\
\noalign{{\color{hlrowcolor}\hrule height 9.3pt}\vskip -9.3pt}
FN-Prune & 74.43 & \textbf{61.43} & 44.60 & \underline{41.40} & \textbf{71.35} & 66.46 & 9.43 & \textbf{10.15} & 81.65 & \textbf{69.52} \\
\midrule
\multicolumn{11}{@{}l}{\textbf{\textit{ERNIE-4.5-21B-A3B-PT}}} \\
Full model & \multicolumn{2}{c}{73.87} & \multicolumn{2}{c}{46.40} & \multicolumn{2}{c}{71.74} & \multicolumn{2}{c}{7.26} & \multicolumn{2}{c}{78.39} \\
\noalign{\vskip 1pt\hbox to \textwidth{\xleaders\hbox{\rule{2.2pt}{0.25pt}\hskip 2pt}\hfill}\vskip 1pt}
M-SMoE~{\fontsize{6.2}{6.8}\selectfont\citeyearpar{li2023mergemsmoe}} & 55.68 & 31.96 & 37.80 & 29.40 & 65.04 & 56.67 & 10.58 & 19.52 & 63.00 & 3.34 \\
MC-SMoE~{\fontsize{6.2}{6.8}\selectfont\citeyearpar{li2023mergemsmoe}} & 56.59 & 36.60 & 40.20 & 29.80 & 65.75 & 56.27 & 9.11 & 15.38 & 63.84 & 3.41 \\
OWL-to-MoE~{\fontsize{6.2}{6.8}\selectfont\citeyearpar{yin2024owl}} & \underline{66.58} & 29.07 & 41.20 & 31.80 & 66.69 & 55.96 & 9.92 & 20.32 & 17.13 & 0.53 \\
NAEE~{\fontsize{6.2}{6.8}\selectfont\citeyearpar{lu2024notallequal}} & 48.78 & 33.21 & 28.60 & 25.60 & 54.14 & 50.43 & 18.82 & 27.16 & 30.55 & 5.99 \\
EEP-Prune~{\fontsize{6.2}{6.8}\selectfont\citeyearpar{liu2024eep}} & \textbf{66.72} & 46.10 & \underline{46.20} & \underline{38.80} & 69.77 & 62.35 & \textbf{7.94} & \underline{12.81} & \textbf{80.14} & \underline{52.99} \\
AlphaPruning-to-MoE~{\fontsize{6.2}{6.8}\selectfont\citeyearpar{lu2024alphapruning}} & 53.70 & 36.02 & 40.00 & 26.40 & 66.30 & 48.38 & 11.32 & 21.19 & 77.71 & 30.86 \\
HC-SMoE~{\fontsize{6.2}{6.8}\selectfont\citeyearpar{chen2025hcsmoe}} & 62.67 & \underline{48.28} & 41.40 & 32.40 & 68.75 & 59.59 & 10.88 & 17.66 & 66.11 & 26.38 \\
MoE Pathfinder~{\fontsize{6.2}{6.8}\selectfont\citeyearpar{yang2025moepathfinder}} & 64.55 & 47.81 & \textbf{46.60} & \textbf{39.60} & \underline{69.93} & \underline{65.51} & 8.28 & 13.70 & \underline{78.24} & 40.79 \\
REAP-Prune~{\fontsize{6.2}{6.8}\selectfont\citeyearpar{lasby2025reap}} & 47.98 & 24.44 & 34.60 & 25.20 & 65.51 & 53.75 & 13.35 & 20.99 & 68.39 & 11.37 \\
MoNE~{\fontsize{6.2}{6.8}\selectfont\citeyearpar{zhang2026mone}} & 53.91 & 32.20 & 37.00 & 29.40 & 67.48 & 56.20 & 11.82 & 19.99 & \textbf{80.14} & 36.32 \\
\noalign{{\color{hlrowcolor}\hrule height 9.3pt}\vskip -9.3pt}
FN-Prune & 61.70 & \textbf{50.46} & 46.00 & 38.40 & \textbf{71.35} & \textbf{66.61} & \underline{8.27} & \textbf{11.94} & \underline{78.24} & \textbf{59.97} \\
\midrule
\multicolumn{11}{@{}l}{\textbf{\textit{OLMoE-1B-7B-0924}}} \\
Full model & \multicolumn{2}{c}{50.39} & \multicolumn{2}{c}{45.40} & \multicolumn{2}{c}{68.67} & \multicolumn{2}{c}{15.41} & \multicolumn{2}{c}{8.11} \\
\noalign{\vskip 1pt\hbox to \textwidth{\xleaders\hbox{\rule{2.2pt}{0.25pt}\hskip 2pt}\hfill}\vskip 1pt}
M-SMoE~{\fontsize{6.2}{6.8}\selectfont\citeyearpar{li2023mergemsmoe}} & 37.24 & 24.24 & 34.20 & 25.60 & 59.27 & \underline{54.54} & 17.53 & 138.80 & 0.68 & 0.00 \\
MC-SMoE~{\fontsize{6.2}{6.8}\selectfont\citeyearpar{li2023mergemsmoe}} & \underline{38.11} & 24.50 & 34.80 & 25.60 & 59.35 & 53.43 & 17.02 & 79.77 & 0.08 & 0.00 \\
OWL-to-MoE~{\fontsize{6.2}{6.8}\selectfont\citeyearpar{yin2024owl}} & 37.74 & 24.90 & 39.80 & 26.20 & 61.56 & 52.64 & 19.47 & 451.60 & 1.67 & 0.00 \\
NAEE~{\fontsize{6.2}{6.8}\selectfont\citeyearpar{lu2024notallequal}} & 22.88 & 24.12 & 23.60 & 25.60 & 50.28 & 51.30 & 394.82 & 3778.51 & 0.00 & 0.00 \\
EEP-Prune~{\fontsize{6.2}{6.8}\selectfont\citeyearpar{liu2024eep}} & 35.38 & 23.16 & 39.60 & 30.60 & 61.56 & 52.88 & \textbf{13.42} & 53.50 & \underline{2.73} & \textbf{0.68} \\
AlphaPruning-to-MoE~{\fontsize{6.2}{6.8}\selectfont\citeyearpar{lu2024alphapruning}} & 23.25 & 23.69 & 26.40 & 24.40 & 51.22 & 51.38 & 104.13 & 3768.09 & 0.30 & 0.00 \\
HC-SMoE~{\fontsize{6.2}{6.8}\selectfont\citeyearpar{chen2025hcsmoe}} & 37.10 & \textbf{26.36} & 36.00 & 27.60 & 63.54 & \textbf{54.78} & 29.30 & 197.20 & 1.82 & 0.00 \\
MoE Pathfinder~{\fontsize{6.2}{6.8}\selectfont\citeyearpar{yang2025moepathfinder}} & 36.95 & \underline{25.94} & \underline{41.00} & \underline{32.40} & \underline{63.69} & \underline{54.54} & 20.00 & \underline{48.03} & 1.90 & \underline{0.61} \\
REAP-Prune~{\fontsize{6.2}{6.8}\selectfont\citeyearpar{lasby2025reap}} & 22.94 & 23.35 & 25.40 & 24.60 & 50.67 & 51.54 & 871.32 & 3595.99 & 0.00 & 0.00 \\
MoNE~{\fontsize{6.2}{6.8}\selectfont\citeyearpar{zhang2026mone}} & 23.91 & 24.76 & 27.40 & 25.00 & 52.09 & 49.17 & 23.68 & 109.30 & 2.05 & 0.00 \\
\noalign{{\color{hlrowcolor}\hrule height 9.3pt}\vskip -9.3pt}
FN-Prune & \textbf{40.69} & 25.34 & \textbf{42.80} & \textbf{33.40} & \textbf{65.27} & 53.20 & \underline{16.19} & \textbf{32.48} & \textbf{3.49} & 0.30 \\
\bottomrule
\end{tabular*}
\par\smallskip
\begin{minipage}{\textwidth}\fontsize{8}{10}\selectfont All entries except WT2 perplexity are accuracies (\%); lower WT2 perplexity is better. The best and second-best compressed models per condition are bold and underlined, with the full model excluded. OBQA = OpenBookQA; WinoG.\ = WinoGrande; WT2 = WikiText-2.\end{minipage}
\end{table*}

\endgroup

\subsection{Experimental Setup}
\label{sec:baseline_protocol}

\paragraph{Models and budgets}
We study Qwen3-30B-A3B~\citep{qwen2025technical}, ERNIE-4.5-21B-A3B-PT~\citep{ernie2025technical}, and OLMoE-1B-7B-0924~\citep{muennighoff2024olmoe}, whose routed-expert pools hold $48{\times}128$, $27{\times}64$, and $16{\times}64$ slots; ERNIE additionally carries two shared experts per layer, which every method preserves. We evaluate removal budgets of $25\%$ and $50\%$ of the routed experts. All methods remove the same number of routed-expert slots while keeping their native compression operation, whether deletion, novice replacement by constant modules, merging, or merging followed by recovery training. FN-Prune scores with the checkpoint's native selection rule, routing weights, and correction bias, and applies the same policy to the surviving experts after deletion.

\paragraph{Calibration and development}
Calibration, development, and token-weighted perplexity use the complete train, validation, and test splits of WikiText-2~\citep{merity2017wikitext}; the tokenized streams are packed at context length $2{,}048$, and the final, possibly partial chunk is retained. Search and recovery methods follow their published procedures.

\paragraph{Task evaluation}
We report zero-shot accuracy on MMLU~\citep{hendrycks2021mmlu} and WinoGrande~\citep{sakaguchi2021winogrande}, character-normalized continuation scores on OpenBookQA~\citep{mihaylov2018openbookqa}, five-shot strict-match accuracy on GSM8K~\citep{cobbe2021gsm8k}, and token-weighted WikiText-2 perplexity. The three-task mean averages MMLU, OpenBookQA, and WinoGrande accuracy. Qwen3 reasoning is disabled and no chat template is applied, and every configuration runs with five seeds on the complete population of each benchmark (MMLU $14{,}042$; OpenBookQA $500$; WinoGrande $1{,}267$; GSM8K $1{,}319$; WikiText-2 test $4{,}358$ source rows). Each value is the average over five random seeds, and the standard deviation across seeds is less than 0.1.

\subsection{RQ1 (Published Baselines)}
\label{sec:rq1_frontier}
\label{sec:baseline_portfolio}

The matched ranking control is frequency deletion, which removes the least frequently selected experts under exactly the quotas and deletion procedure that FN-Prune uses. The published comparison spans ten methods, namely the allocation methods OWL-to-MoE~\citep{yin2024owl} and AlphaPruning-to-MoE~\citep{lu2024alphapruning}, the deletion methods NAEE~\citep{lu2024notallequal}, EEP-Prune~\citep{liu2024eep}, MoE Pathfinder~\citep{yang2025moepathfinder}, and REAP-Prune~\citep{lasby2025reap}, the novice replacement MoNE~\citep{zhang2026mone}, and the merging systems M-SMoE, MC-SMoE~\citep{li2023mergemsmoe}, and HC-SMoE~\citep{chen2025hcsmoe}. Each method keeps its own operation at the same total routed-expert budget. The frontier denotes the strongest published score in each condition, and the unpruned checkpoints serve as references; implementation adaptations appear in Appendix~\ref{app:baseline_implementations}.
\label{sec:frontier}

\begin{figure*}[ht]
    \centering
    \includegraphics[width=0.8\textwidth]{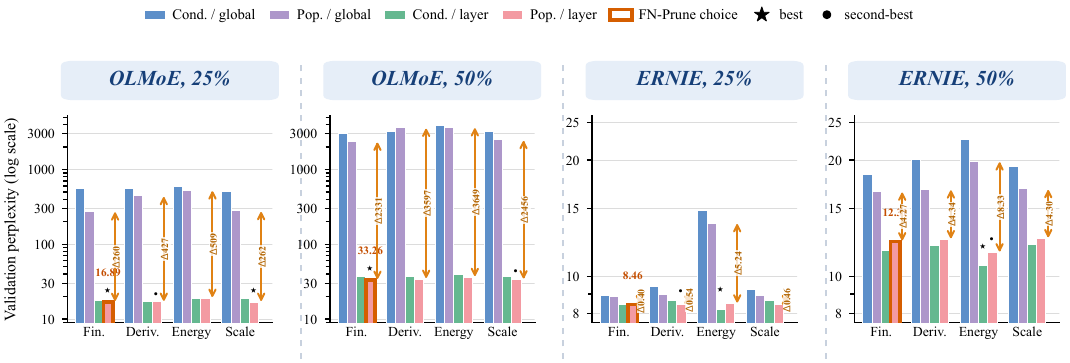}
    \caption{\textbf{Factorial ablation of score, aggregation, and allocation.} WikiText-2 validation perplexity after physical deletion, one panel per model--budget condition (logarithmic axes; lower is better). Arrows mark the reduction obtained by switching from population/global to population/layer allocation.}
    \label{fig:design_space_controls}
\end{figure*}

Across all five metrics and six conditions (Table~\ref{tab:baseline_results}), removing half the experts lowers WikiText-2 perplexity on every family. The margin is widest on OLMoE, where FN-Prune reaches $32.48$ against a frontier of $48.03$, a $32\%$ reduction. This is also the family where deletion hurts most, because every token still activates $k_\ell=8$ experts and half of the original pool of $64$ must absorb all routing demand. The picture reverses at $25\%$ removal, where the published frontier keeps its perplexity advantage and MC-SMoE leads on Qwen3. MC-SMoE is the one portfolio method with recovery training, and its edge fades exactly where rankings start to dominate. Beyond perplexity, FN-Prune attains the best compressed-model score more often than any published method, and the frontier is spread across MoE Pathfinder, EEP-Prune, MC-SMoE, and HC-SMoE, so no single method dominates. The closest competitor at the aggressive budget, EEP-Prune, is itself a deletion method, suggesting that with the compression operation fixed, the comparison comes down to ranking quality.

\begin{figure*}[ht]
    \centering
    \includegraphics[width=0.8\textwidth]{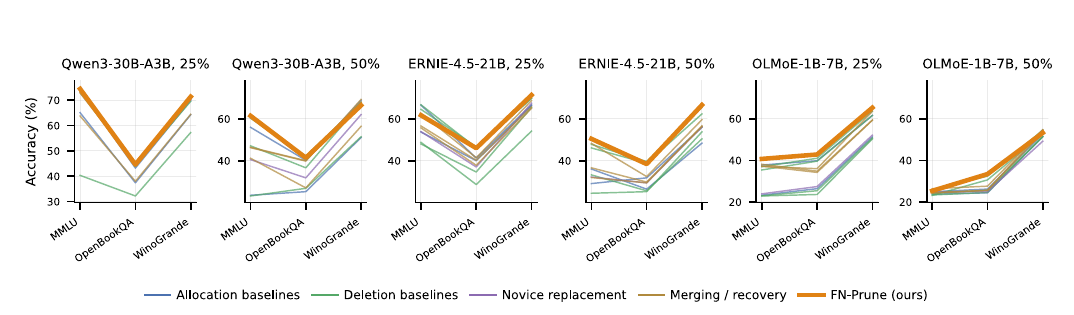}
    \caption{\textbf{Per-task accuracy profiles (parallel coordinates).} One polyline per method across the three decision tasks on a shared percent scale; one panel per family and removal budget. FN-Prune is bold orange; published baselines are thin lines colored by method group; all values are complete-population evaluations.}
    \label{fig:task_profiles}
\end{figure*}

% Generated by tools/render_baseline_paper_tables_v2.py; do not edit numeric rows.
% Source: evidence/baseline_results_final_96_20261005/baseline_campaign_audit.json
\begin{table*}[ht]
\centering
\caption{Generation throughput (generated tokens/s) for the ten published methods and FN-Prune (shaded) at 25\% and 50\% removal; the full-model row is the unpruned reference.}
\label{tab:baseline_efficiency}
\fontsize{7.5}{8.8}\selectfont
\setlength{\tabcolsep}{1.5pt}
\renewcommand{\arraystretch}{1.0}
\begin{tabular*}{\textwidth}{@{\extracolsep{\fill}}lrrrrrr@{}}
\toprule
& \multicolumn{2}{c}{Qwen3-30B-A3B} & \multicolumn{2}{c}{ERNIE-4.5-21B-A3B-PT} & \multicolumn{2}{c}{OLMoE-1B-7B-0924} \\
\cmidrule(lr){2-3}\cmidrule(lr){4-5}\cmidrule(lr){6-7}
Method & 25\% & 50\% & 25\% & 50\% & 25\% & 50\% \\
\midrule
Full model & \multicolumn{2}{c}{291.84} & \multicolumn{2}{c}{2043.80} & \multicolumn{2}{c}{511.94} \\
\noalign{\vskip 1pt\hbox to \textwidth{\xleaders\hbox{\rule{2.2pt}{0.25pt}\hskip 2pt}\hfill}\vskip 1pt}
M-SMoE~{\fontsize{6.2}{6.8}\selectfont\citeyearpar{li2023mergemsmoe}} & 345.45 & 484.71 & 2455.52 & \textbf{3307.64} & 239.24 & 960.00 \\
MC-SMoE~{\fontsize{6.2}{6.8}\selectfont\citeyearpar{li2023mergemsmoe}} & 339.42 & 455.09 & \textbf{2611.54} & 3275.09 & 676.92 & 977.37 \\
OWL-to-MoE~{\fontsize{6.2}{6.8}\selectfont\citeyearpar{yin2024owl}} & 352.20 & 495.26 & 2583.34 & 3211.66 & 619.82 & 495.17 \\
NAEE~{\fontsize{6.2}{6.8}\selectfont\citeyearpar{lu2024notallequal}} & 334.32 & 508.25 & \underline{2599.61} & 3216.63 & \underline{700.14} & \textbf{1062.97} \\
EEP-Prune~{\fontsize{6.2}{6.8}\selectfont\citeyearpar{liu2024eep}} & 355.75 & 466.78 & 2421.61 & 3284.98 & \textbf{706.04} & 972.38 \\
AlphaPruning-to-MoE~{\fontsize{6.2}{6.8}\selectfont\citeyearpar{lu2024alphapruning}} & \underline{373.10} & \textbf{897.46} & 2536.41 & \underline{3298.94} & 353.49 & 451.34 \\
HC-SMoE~{\fontsize{6.2}{6.8}\selectfont\citeyearpar{chen2025hcsmoe}} & 371.26 & 562.43 & 2550.78 & 3265.20 & 351.35 & 934.16 \\
MoE Pathfinder~{\fontsize{6.2}{6.8}\selectfont\citeyearpar{yang2025moepathfinder}} & 333.08 & 489.02 & 2458.61 & 3287.91 & 662.75 & 952.93 \\
REAP-Prune~{\fontsize{6.2}{6.8}\selectfont\citeyearpar{lasby2025reap}} & \textbf{509.45} & \underline{827.53} & 2424.30 & 3252.66 & 335.29 & \underline{1059.64} \\
MoNE~{\fontsize{6.2}{6.8}\selectfont\citeyearpar{zhang2026mone}} & 314.79 & 593.61 & 2212.84 & 2429.68 & 611.94 & 655.86 \\
\noalign{{\color{hlrowcolor}\hrule height 9.2pt}\vskip -9.2pt}
FN-Prune & 346.87 & 481.74 & 2441.15 & 2909.16 & 630.98 & 910.19 \\
\bottomrule
\end{tabular*}
\end{table*}

\subsection{RQ2 (Design Choices)}
\label{sec:rq2_ablation}
\label{sec:results_headline}
\label{sec:method_controls}

We ablate the statistic, the aggregation rule, and the allocation policy on the complete WikiText-2 validation split. Sixteen configurations cross four statistics (routed energy $\|c_i(t)\|^2$, the local derivative in Eq.~\ref{eq:rmsnorm_jvp}, the finite difference in Eq.~\ref{eq:fn_score}, and the scale-only control $\|\gamma\odot c_i(t)\|^2/v$ with $v=\tfrac1d\|r_t\|^2+\epsilon$) with conditional or population averaging and with uniform-layer or global allocation; conditional averaging uses the tokens that select an expert, population averaging all calibration tokens, and the global allocator retains per-layer top-$k$ capacity. Every configuration is evaluated after physical deletion. The resulting grid (Figure~\ref{fig:design_space_controls}) covers OLMoE-1B-7B and ERNIE-4.5-21B at both budgets; Qwen3 was reserved for the formal evaluation, and the selected configuration is applied unchanged to all six.

Allocation dominates the ablation. Global quotas drive OLMoE perplexity above $2{,}300$ at $50\%$ removal, whereas the adopted layer allocation stays in the tens, and within layer allocation the finite difference with population averaging achieves the lowest mean over the four conditions. Conditional routed energy is stronger on ERNIE, but the selected rule is strongest on OLMoE and gives the best overall mean. The ordering of the statistics is consistent with the decomposition in Eq.~\ref{eq:fn_decomposition}. Each partial statistic omits one of these ingredients, whether the cancellation between the content and radial terms, the second-order scale change, or the residual context; only the full finite difference keeps all three. Population aggregation is justified by the identity in Section~\ref{sec:method_score}, since a rarely selected expert with a modest conditional disturbance is still cheap to remove and only the population mean captures this.

\label{sec:method_motivation}
Preliminary OLMoE measurements show how strongly allocation alone moves perplexity. Perturbing $30\%$ of one layer's routing decisions changes perplexity by several percent, most strongly in the deepest layers (Spearman $r_s=0.576$, $p=0.019$ for correlation with depth); at a nominal $5\%$ budget, uniform removal of $48$ experts raises perplexity from $11.82$ to $496.85$, while the same $48$ removals allocated non-uniformly reach $27.98$. We therefore hold per-layer quotas fixed when isolating the ranking criterion.

The matched-quota control changes which experts leave each layer while holding the number leaving it fixed, testing whether the ranking alone improves the resulting model. With quotas and deletion held fixed, FN-Prune lowers WikiText-2 perplexity at $50\%$ removal on all three families, most strongly on OLMoE, where perplexity falls from $49.14$ to $32.48$. At $25\%$ removal the ranking matters less, and perplexity improves only on ERNIE, from $8.50$ to $8.27$. The budget asymmetry is intuitive. When only a quarter of the experts are removed, both rankings avoid most harmful ones and the surviving pools hardly differ; at half removal the pool is at its minimum, and small ranking differences compound layer by layer.

\subsection{RQ3 (Capability Retention)}
\label{sec:rq3_retention}

Across the three decision tasks on a common percentage scale (Figure~\ref{fig:task_profiles}), perplexity and downstream retention favor different conditions. At $50\%$ removal, FN-Prune improves GSM8K over the strongest published method by $7.50$ and $6.98$ percentage points on Qwen3 and ERNIE. This transfer is expected from how the criterion is built, since five-shot greedy generation rewards the language-modeling fidelity that the score measures, while multiple-choice discrimination depends on features it does not. FN-Prune on OLMoE instead leads GSM8K at $25\%$ removal, with absolute accuracies of $3.49\%$ and $0.30\%$; the unpruned checkpoint itself scores only $8.11\%$, so arithmetic generation is inherently fragile in this family.

Multiple-choice results are similarly selective. FN-Prune improves WinoGrande on every family (both budgets on ERNIE), MMLU on Qwen3 and ERNIE at $50\%$, and OpenBookQA on OLMoE at both budgets; on Qwen3, WinoGrande is the only task that improves, and the ERNIE advantage widens with the removal budget.

To examine balance across decision tasks, we pair their equal-weight mean accuracy with the population standard deviation of their changes from the unpruned model; a model is non-dominated among the eleven compressed models when no competitor is at least as good on both and strictly better on one. EEP-Prune dominates ERNIE at $25\%$ and Qwen3 at $50\%$, and FN-Prune is non-dominated in every other condition while typically holding the highest three-task mean. The method thus concentrates its advantage where language modeling matters most and concedes scattered multiple-choice conditions to systems with search or recovery training. The unpruned references exceed the published frontier almost everywhere, with the exceptions confined to $25\%$ removal; FN-Prune thus improves retention over the compressed alternatives while remaining below the unpruned checkpoints. Its configuration was selected with WikiText-2 validation perplexity and no task labels.

\subsection{RQ4 (Physical Efficiency)}
\label{sec:rq4_efficiency}
\label{sec:efficiency}

Profiling each configuration once on an A100, we measure generation throughput over the materialized models, which are reconstructed from shared checkpoints and compression recipes or adapters. Because novice replacement swaps full expert modules for learned constants, it changes the footprint only marginally, by at most $0.024$ GiB. At $50\%$ removal the materialized footprint roughly halves on all three families, from $56.87$ to $29.86$ GiB on Qwen3, from $40.66$ to $21.67$ GiB on ERNIE, and from $12.89$ to $6.89$ GiB on OLMoE. The reduction is slightly below one half because attention weights and, in ERNIE, the always-on shared experts remain. This halving comes from deleting the weights physically, and a mask over the same experts would change neither footprint nor traffic.

FN-Prune's measured throughput rises with the removal budget on all three families (the shaded row in Table~\ref{tab:baseline_efficiency}), plausibly through smaller routing computations and reduced memory traffic over a lighter model; on OLMoE, generation moves from $512$ tokens per second in the unpruned model to $631$ and then $910$ as the budget grows. Throughput tracks the slot budget far more than the ranking, since every method at a given budget activates the same number of experts per token and the residual spread across methods reflects implementation factors. Since batches and device conditions differ across families, these measurements support within-family trends, and serving latency warrants its own study.
\section{Conclusion and Limitations}
\label{sec:conclusion}
\label{sec:limitations}

FN-Prune turns the finite normalized effect of removing one routed contribution into a one-pass criterion for physical expert deletion. At $50\%$ removal, it improves WikiText-2 perplexity across all three families and GSM8K on Qwen3 and ERNIE relative to the strongest published methods. The task profiles locate its main benefit in aggressive language-model compression, with selective downstream retention.

Our study also has several limitations. We evaluate three MoE families under two removal budgets, with calibration drawn from WikiText-2 and uniform per-layer quotas. Experts are scored under the checkpoint’s native routing, whereas deployment reroutes among the survivors; the score is therefore a fixed-routing proxy for the pruned model. We do not combine FN-Prune with recovery training or study non-uniform layer allocation, and physical deletion assumes that expert modules and router rows can be sliced cleanly while shared experts are retained. Extending FN-Prune to deployment-aware scoring, adaptive quotas, and post-pruning recovery is a natural next step.
% Main body ends here. End matter starts on fresh pages: ICML 2027 excludes the
% impact statement, AI disclosure, references, and appendices from the page limit.
\clearpage
\section*{Impact Statement}
Reducing model memory can make existing language models more accessible, but it does not establish that their outputs are more reliable. Our results show that pruning can improve perplexity relative to other compressed models while downstream retention remains task-dependent. Compressed models should therefore be evaluated on their intended tasks before deployment, especially where errors have consequential effects. We do not assess safety, fairness, or real-world reliability, and the compression results should not be interpreted as evidence for those properties.

\section*{AI Disclosure Statement}
Substantial assistance from generative AI tools was used in conducting this research and preparing this manuscript, including research ideation, experimental design, code generation for the scoring, pruning, and evaluation pipelines, data analysis, and manuscript drafting. All reported results were produced by the deterministic, seed-fixed pipelines described in the paper under one frozen method version and were cross-checked against frozen experimental receipts. All scientific decisions and final claims were made by the authors, who remain fully responsible for the correctness, originality, and integrity of the content.
\bibliographystyle{icml2027}
\bibliography{references}
\appendix
\section{Proofs}
\label{app:proofs}

\paragraph{Consumer-side decomposition (Proposition~\ref{prop:decomposition}).}
\begin{proof}
Write $r/\sqrt v-(r-c)/\sqrt{v'}=c/\sqrt{v'}+r(\sqrt{v'}-\sqrt v)/(\sqrt v\sqrt{v'})$. Substituting $\sqrt{v'}-\sqrt v=(v'-v)/(\sqrt v+\sqrt{v'})$ and $v'-v=(\|c\|^2-2\langle r,c\rangle)/d$, then multiplying channelwise by $\gamma$, gives Eq.~\ref{eq:fn_decomposition}.
\end{proof}

\paragraph{Survivor renormalization (Proposition~\ref{prop:refill}).}
\begin{proof}
The renormalized survivor output is $y'=(y_\ell-w_i f_i)/(1-w_i)$. Hence $\Delta r_t=y'-y_\ell=w_i(y_\ell-f_i)/(1-w_i)=(w_i y_\ell-c_i)/(1-w_i)$ because $c_i=w_i f_i$.
\end{proof}

\paragraph{One-pass accumulation (Proposition~\ref{prop:onepass}).}
\begin{proof}
Equation~\ref{eq:fn_decomposition} is the finite difference in Eq.~\ref{eq:fn_score}. Algorithm~\ref{alg:fnprune} observes and accumulates it for each selected pair $(r_t,c_i(t))$, while unselected pairs have $c_i(t)=0$. Evaluating $k_\ell$ differences per token costs $O(k_\ell d)$ arithmetic, and one scalar per expert costs $O(\sum_\ell N_\ell)$ storage.
\end{proof}

\section{Baseline Implementations}
\label{app:baseline_implementations}

The main table uses the ten published methods named in Section~\ref{sec:baseline_portfolio}. NAEE uses the additive reconstruction proxy, while OWL uses its available direction-based allocation adaptation. EEP-Prune searches physical-deletion plans with population $20$ for $40$ iterations on complete validation loss, and REAP-Prune collects its saliency with router-weight renormalization before physical deletion.

MC-SMoE applies one epoch of last-hidden-state distillation on the complete WikiText-2 training split, using rank-$32$ LoRA on retained routed experts; the other methods retain their stated deletion, replacement, or merging operation. These adaptations affect the final compressed models; the frequency control keeps FN-Prune's quotas and deletion fixed, isolating the ranking criterion.

\end{document}